\documentclass[journal]{IEEEtran}
\usepackage[font=small,skip=1pt]{caption}
\IEEEoverridecommandlockouts
\usepackage{textcomp}
\usepackage{gensymb}
\usepackage{latexsym}
\usepackage[T1]{fontenc}
\usepackage{multirow}
\usepackage{graphicx}
\usepackage{varioref}
\usepackage{array}
\usepackage{enumitem}
\usepackage{soul,color}
\usepackage{longtable}
\usepackage{pifont}
\usepackage[noend]{algpseudocode}
\usepackage{booktabs}

\usepackage[utf8]{inputenc} 
\usepackage[T1]{fontenc}    
\usepackage{url}            
\usepackage{booktabs}       
\usepackage{amsfonts}       
\usepackage{nicefrac}       
\usepackage{microtype}      
\usepackage{xcolor}         
\usepackage{graphicx}
\usepackage{caption}
\usepackage{subcaption}
\usepackage[table,xcdraw]{xcolor}

\usepackage{algorithm}
\usepackage{float}

\usepackage{color}
\definecolor{light}{rgb}{0.5, 0.5, 0.5}

\usepackage{graphicx}
\medskip
\usepackage{amssymb}
\usepackage{amsmath,epsfig,amssymb}
\usepackage{amsthm}
\usepackage[active]{srcltx} 
\usepackage{epstopdf} 

\usepackage{bm}
\usepackage{enumitem}
\DeclareRobustCommand*{\IEEEauthorrefmark}[1]{%
  \raisebox{0pt}[0pt][0pt]{\textsuperscript{\footnotesize\ensuremath{#1}}}}

\usepackage{cite}
\usepackage{amsmath,amssymb,amsfonts}
\usepackage{graphicx}
\usepackage{textcomp}
\usepackage{xcolor}
\def\BibTeX{{\rm B\kern-.05em{\sc i\kern-.025em b}\kern-.08em
    T\kern-.1667em\lower.7ex\hbox{E}\kern-.125emX}}
\usepackage[english]{babel}
\usepackage{mathdots}

\graphicspath{{Pictures/}{../jpeg/}} 
\usepackage{float}
\usepackage{amsmath}
\usepackage{hyperref}

\begin{document}

\title{EEG-to-Text Translation: A Model for Deciphering Human Brain Activity}

\author{\IEEEauthorblockN{Saydul Akbar Murad\IEEEauthorrefmark{1}, Ashim Dahal\IEEEauthorrefmark{1}
Nick Rahimi\IEEEauthorrefmark{1}}

\thanks{\IEEEauthorrefmark{1} School of Computing Sciences \& Computer Engineering, University of Southern Mississippi, Hattiesburg, MS 39406, USA. (e-mail: saydulakbar.murad@usm.edu, ashim.dahal@usm.edu, nick.rahimi@usm.edu)}
}  




\maketitle
\begin{abstract}
With the rapid advancement of large language models like Gemini, GPT, and others, bridging the gap between the human brain and language processing has become an important area of focus. To address this challenge, researchers have developed various models to decode EEG signals into text. However, these models still face significant performance limitations. To overcome these shortcomings, we propose a new model, R1 Translator, which aims to improve the performance of EEG-to-text decoding. The R1 Translator model combines a bidirectional LSTM encoder with a pretrained transformer-based decoder, utilizing EEG features to produce high-quality text outputs. The model processes EEG embeddings through the LSTM to capture sequential dependencies, which are then fed into the transformer decoder for effective text generation. The R1 Translator excels in ROUGE metrics, outperforming both T5 (previous research) and Brain Translator. Specifically, R1 achieves a ROUGE-1 score of 38.00\% (P), which is up to 9\% higher than T5 (34.89\%) and 3\% better than Brain (35.69\%). It also leads in ROUGE-L, with a F1 score of 32.51\%, outperforming T5 by 3\% (29.67\%) and Brain by 2\% (30.38\%). In terms of CER, R1 achieves a CER of 0.5795, which is 2\% lower than T5 (0.5917) and 4\% lower than Brain (0.6001). Additionally, R1 performs better in WER with a score of 0.7280, outperforming T5 by 4.3\% (0.7610) and Brain by 3.6\% (0.7553). The code for this research can be found in this link: \href{https://github.com/Mmurrad/EEG-To-text}{https://github.com/Mmurrad/EEG-To-text}
\end{abstract} 

\begin{table}[h]
\centering
\caption{Evaluation of Ground Truth and R1 Translator Output for EEG-to-Text Decoding}
\label{comparison_table}
\resizebox{\columnwidth}{!}{%
\begin{tabular}{@{}ll@{}}
\toprule
Ground Truth  & Fans \textbf{of the} TV \textbf{series will be} disappointed, and everyone else \textbf{will be} slightly bored. \\ \midrule
R1 Translator & \textbf{of the} film \textbf{series will be} familiar to but the will \textbf{will be} happy disappointed.       \\ \bottomrule
\end{tabular}%
}
\end{table}

\section{Introduction}
\label{intro}

Decoding human cognition into text has garnered significant attention from researchers due to its profound implications. Among various techniques, Electroencephalogram (EEG) signals are particularly favored because of their non-invasive nature and ease of acquisition. Traditional EEG decoding approaches primarily focus on classifying brain states into specific categories, such as Motor Imagery (MI) \cite{intro1}, emotions \cite{intro2, IA3},  attention \cite{intro3}, mental workload \cite{intor4}, and cognitive states \cite{intro5}. However, limiting decoding to mere classification is insufficient for the broader ambitions within brain-computer communication. Recent advancements in large language models (LLMs) have opened new avenues for bridging the gap between natural language processing and brain signals, presenting exciting possibilities for more comprehensive and intuitive brain-computer interfaces.

Despite recent advances in EEG to text translation \cite{intro6, intro7, intro8, intro9}, several key issues remain, particularly concerning the robustness and scalability of current models. A major concern in this field is the use of implicit teacher forcing (TF) during model evaluation \cite{intro9}. This technique can inadvertently inflate performance metrics, creating an unrealistic representation of the true capabilities of the model. Although many researchers have investigated the differences between teacher-forcing and non-teacher-forcing approaches, these studies still face limitations in terms of performance and generalizability \cite{intro6}. A crucial aspect of this comparison is the lack of testing against pure noise; similar performance on noise would suggest that the model is not learning effectively from EEG signals but is instead memorizing text \cite{intro8}. Furthermore, much of the existing research has been restricted to small, often non-representative datasets, which fail to accurately reflect the model’s performance in real-world, diverse scenarios \cite{intro10, eeg_review}. As a result, these factors, when overlooked, undermine the validity and reliability of the reported results, leading to an overestimation of the true capabilities of the model.

In this paper, we introduce R1 Translator, a novel architecture framework designed to turn EEG signals into text. The model works in two stages: first, it uses an LSTM encoder to process the raw EEG data, and then it projects these features into a decoder embedding space. This method utilizes the power of a pretrained BART model \cite{BART} for generating text, making sure the output is fluent and contextually accurate. By using a bidirectional LSTM \cite{Bi-LSTM}, the model can consider both past and future information from the EEG signals, capturing the full sequence of brainwave patterns. Unlike many previous methods, R1 Translator doesn’t rely on external markers like eye-tracking, which can be cumbersome and imprecise. Instead, it directly translates the EEG signals into text, offering a more practical solution for real-world applications. The combination of a powerful pretrained language model with a custom EEG encoder allows R1 Translator to generalize better and make the EEG-to-text process more interpretable. 

In this research experiment, we employed non-invasive EEG signals from the ZuCo dataset (versions 1 \cite{ZuCov1} and 2 \cite{ZuCov2}) to evaluate the performance of three different models: R1 Translator (our proposed model), T5 Translator \cite{intro6}, and Brain Translator \cite{intro6}. The experiment was conducted in three stages: first using version 1 of the dataset, then using version 2, and finally by combining both versions.  R1 Translator outperforms T5 Translator and Brain Translator in key translation metrics, including ROUGE, SacreBLEU, WER, and CER, with a notable advantage of approximately 4-5\% higher performance in BLEU-N and ROUGE scores. The results demonstrate that R1 Translator consistently outperforms the other models across all metrics, achieving state-of-the-art (SOTA) performance in EEG-to-text translation.

\begin{itemize}
    \item We propose R1 Translator, a novel EEG-to-text translation model that utilizes LSTM-based encoding and pretrained language models for superior performance in translating EEG signals to text.

    \item Our experiments demonstrate that R1 Translator achieves SOTA in EEG-to-text translation, outperforming existing models. 

    \item We conduct a thorough evaluation using the ZuCo dataset, testing on multiple versions and combining them to provide a robust performance assessment of R1 Translator.
    
\end{itemize}

The remainder of this paper is organized as follows. Section \ref{rl} reviews the existing literature on EEG-to-text decoding, summarizing recent progress and identifying key methodological gaps that motivate our work. Section \ref{method} details our proposed methodology, presenting the novel R1 Translator architecture and describing the dataset utilized in our experiments. In Section \ref{implementation}, we provide the specific implementation details of our setup. Section \ref{results} presents the experimental results, offering a comprehensive performance analysis of the R1 Translator and comparing our findings against previous research. We then acknowledge the limitations of the current study in Section \ref{limitations}. Finally, Section \ref{conclusion} concludes the paper by summarizing our key contributions and their implications for the field of brain-computer interfaces and also discusses future research directions.

\section{Related Work}
\label{rl}
Over the past few years, a significant amount of research has been dedicated to EEG to text decoding \cite{intro6, intro7, intro8, intro9}. This area of study has attracted attention due to the potential for EEG-based communication systems, especially for individuals with severe motor impairments. Among the various datasets used in this domain, the ZuCo dataset (both versions 1 \cite{ZuCov1} and 2 \cite{ZuCov2}) has become one of the most widely utilized resources for EEG-to-text decoding research. Some researchers have focused primarily on improving the decoding process, attempting to generate highly accurate text from EEG data \cite{intro6, intro7, intro8, intro9}. Others, however, have concentrated on signal classification methods, where the goal is to classify different brain states or emotions based on EEG readings \cite{class1, class2, class3, IA1}.

Earlier research efforts in this domain primarily relied on traditional machine learning (ML) \cite{classML1} and deep learning (DL) \cite{class1} methods, such as Support Vector Machines (SVM) \cite{SVM1} and Convolutional Neural Networks (CNN) \cite{class2, IA2}. These approaches were commonly used for single-label classification tasks, where the goal was to categorize EEG signals into predefined classes. While these models were quite effective for classification purposes, they had inherent limitations, particularly when applied to the more complex task of EEG-to-text decoding. The models' performance was often restricted by the simplicity of the underlying methods, which were unable to capture the full complexity of the EEG signals required for accurate text generation.

In one such study, as outlined in \cite{intro6}, the authors identified a significant flaw in the code used for text decoding in earlier research. Specifically, they discovered a critical error in the baseline code \cite{intro9}, which had been widely used as a reference for EEG-to-text decoding experiments. This error had a substantial impact on the results of previous studies, leading to inaccurate conclusions about the effectiveness of certain decoding methods.

Acknowledging the limitations of previous work, our research sought to address these issues by updating and refining the existing codebase. We worked extensively on improving the decoding process, ensuring that our experiments were conducted with more robust and accurate methodologies.

\section{Methodology}
\label{method}

\begin{figure*}
    \centering
    \includegraphics[width=0.95\linewidth]{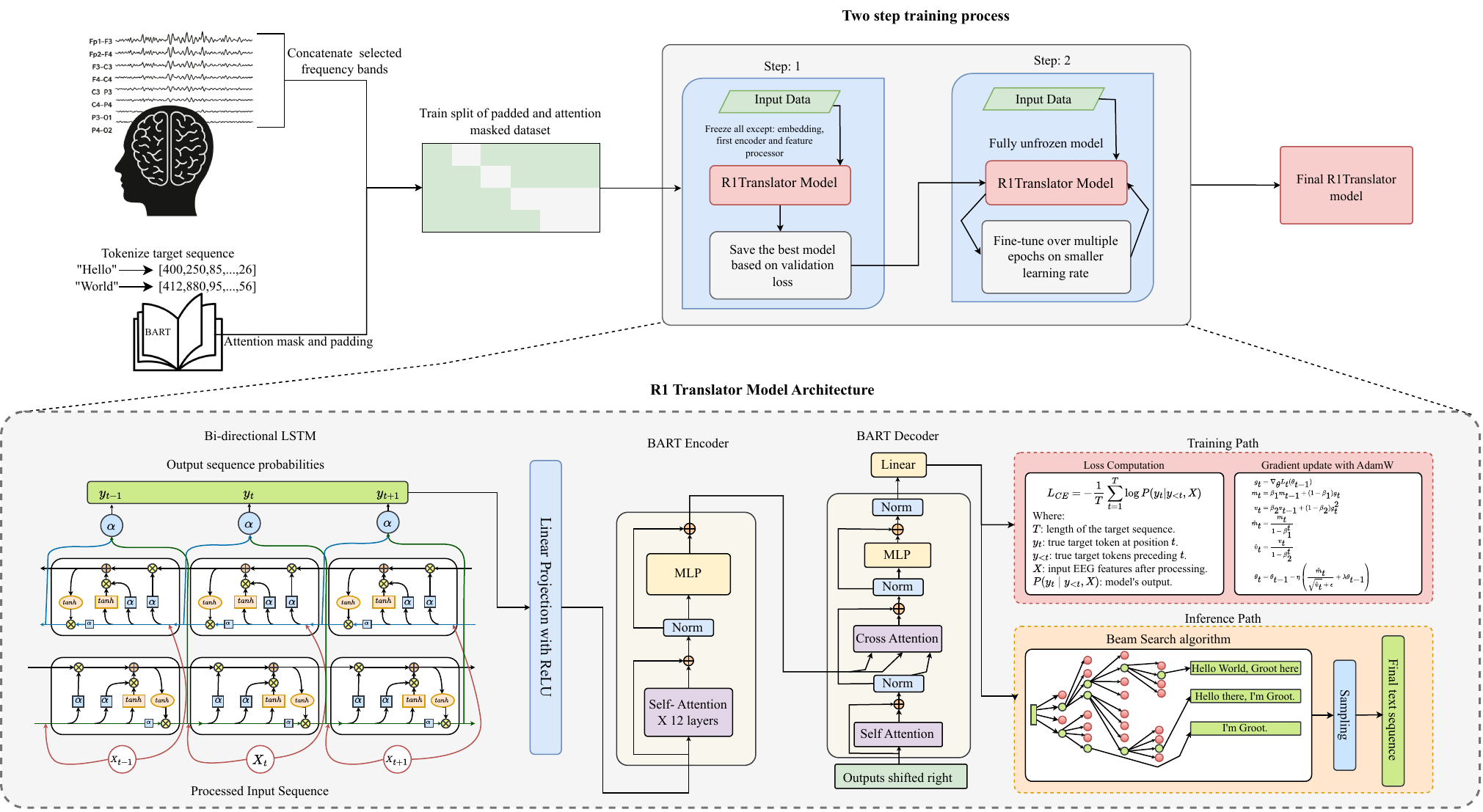}
    \caption{R1Translator model architecture and training principle.}
    \label{methodology}
\end{figure*}

Figure \ref{methodology} provides an overview of the EEG-to-text decoding process, outlining each crucial step. It starts with combining selected frequency bands, followed by the application of attention masking and padding to the tokenized target sequences. The process then moves to a two-step training approach with the R1 Translator, where the first step freezes parts of the model for initial training, and the second step fine-tunes it with a smaller learning rate over several epochs. In the final phase, the model generates the output sequence using a beam search algorithm, producing the decoded text from EEG signals.

\subsection{ZuCo datasets}
In our experiment, we utilized two publicly available ZuCo datasets. These datasets contain EEG signals and corresponding eye-tracking data collected while participants read sentences from movie reviews and Wikipedia articles. The data processing involved segmenting EEG signals based on eye fixations, with each word being assigned a 105-dimensional feature vector derived from eight frequency bands—theta, alpha, beta, and gamma. We focused on the frequency ranges from 4-49.5 Hz for feature extraction. Sentences with any missing or invalid values were excluded from the analysis to ensure data quality. For the dataset split, we divided the data into training (80\%), development (10\%), and test (10\%) sets, ensuring no overlap of sentences across these partitions. We completed the experiment in three different ways. First, we combined the ZuCo V1 dataset, including SR v1.0, NR v1.0, and TSR v1.0. Secondly, we combined ZuCo V1 TSR v2.0 and NR v2.0. Finally, for the third experiment, we combined all datasets from both ZuCo V1 and V2. 

\subsection{R1 Translator Architecture}

The R1 Translator architecture, illustrated in the highlighted portion of Figure \ref{methodology}, is an encoder-decoder framework designed to bridge the gap between high-dimensional neural signals and structured natural language. Its core principle is to synergize the temporal feature extraction capabilities of a Recurrent Neural Network (RNN) with the powerful generative capacity of a pretrained transformer. The architecture consists of three sequential modules: an EEG encoder, a linear projection layer, and a text generation decoder. The complete data flow through these modules is formally detailed in Algorithm~\ref{alg:r1translator}.

\subsubsection{EEG Encoder}
The first module is responsible for processing the sequence of input EEG features, defined in Equation~\eqref{eq:eeg_input}, where each $x_t \in \mathbb{R}^{840}$ is a feature vector for a single word.
\begin{equation}
    X_{EEG} = (x_1, x_2, \ldots, x_T)
    \label{eq:eeg_input}
\end{equation}
To capture both past and future contextual dependencies within this sequence, we employ a multi-layer Bidirectional Long Short-Term Memory (Bi-LSTM) network.

The network processes the input sequence in two directions. The forward LSTM computes a sequence of hidden states, denoted as $\overrightarrow{H}$ in Equation~\eqref{eq:forward_h}, based on the history of the input.
\begin{equation}
    \overrightarrow{H} = (\overrightarrow{h_1}, \overrightarrow{h_2}, \ldots, \overrightarrow{h_T})
    \label{eq:forward_h}
\end{equation}
Concurrently, the backward LSTM computes a sequence of hidden states, $\overleftarrow{H}$ as defined in Equation~\eqref{eq:backward_h}, based on the future of the input.
\begin{equation}
    \overleftarrow{H} = (\overleftarrow{h_1}, \overleftarrow{h_2}, \ldots, \overleftarrow{h_T})
    \label{eq:backward_h}
\end{equation}
At each time step $t$, the final hidden state representation, $h_t$, is formed by concatenating the forward and backward states, as shown in Equation~\eqref{eq:bilstm_concat}.
\begin{equation}
    h_t = [\overrightarrow{h_t} ; \overleftarrow{h_t}]
    \label{eq:bilstm_concat}
\end{equation}
The output of this encoder module is the complete sequence of contextualized EEG representations, $H_{LSTM}$ as shown in Equation~\eqref{eq:lstm_output}, where each $h_t \in \mathbb{R}^{2 \times d_{hidden}}$.
\begin{equation}
    H_{LSTM} = (h_1, h_2, \ldots, h_T)
    \label{eq:lstm_output}
\end{equation}

\subsubsection{Linear Projection}
The feature space of the EEG encoder and the input space of the pretrained language model are dimensionally incompatible. To resolve this, we introduce a linear projection layer that acts as a bridge between the two modules. Each hidden state $h_t$ from the Bi-LSTM is passed through a fully connected layer followed by a Rectified Linear Unit (ReLU) activation function. This transformation is defined by Equation~\eqref{eq:projection}.
\begin{equation}
    z_t = \text{ReLU}(W_{proj}h_t + b_{proj})
    \label{eq:projection}
\end{equation}
Here, $W_{proj}$ and $b_{proj}$ are the trainable weight and bias parameters that project $h_t$ into a new vector $z_t \in \mathbb{R}^{d_{BART}}$, where $d_{BART}$ matches the input embedding dimension of the subsequent BART model. The resulting sequence, $Z = (z_1, z_2, \ldots, z_T)$, serves as the final, aligned representation of the input EEG signals.

\subsubsection{BART-based Text Generation}
The core of our model's translation capability is a pre-trained BART (Bidirectional and Auto-Regressive Transformers) model, which maps the sequence of aligned neural features, $Z$, to a final text sequence, $Y$. This is accomplished through its sophisticated encoder-decoder architecture, built upon the transformer framework. 

\paragraph{BART Encoder}
The processed EEG feature sequence $Z = (z_1, z_2, \ldots, z_T)$ is fed directly as input embeddings to the BART encoder. The encoder's task is to create a rich, context-aware representation of this entire sequence. It consists of a stack of $N$ identical layers, where each layer has two main sub-components: a multi-head self-attention mechanism and a position-wise feed-forward network.

For the self-attention mechanism, the input sequence $Z$ is linearly projected into three matrices: the Query ($Q$), Key ($K$), and Value ($V$). The attention scores are then computed as shown in Equation~\eqref{eq:attention}.
\begin{equation}
    \text{Attention}(Q, K, V) = \text{softmax}\left(\frac{QK^T}{\sqrt{d_k}}\right)V
    \label{eq:attention}
\end{equation}
Here, $d_k$ is the dimension of the keys, used for scaling. This mechanism allows each element in the sequence to attend to all other elements, weighing their importance to generate a new representation. Multi-head attention enhances this by performing the attention mechanism multiple times in parallel with different linear projections. The output of the attention sub-layer is then passed through a feed-forward network. The final output of the encoder stack is a sequence of context-rich hidden states, $H_{enc}$.

\paragraph{BART Decoder}
The BART decoder is an auto-regressive model that generates the text sequence $Y = (y_1, y_2, \ldots, y_{L})$ one token at a time. Like the encoder, it is a stack of $N$ identical layers, but each layer contains three sub-components.

First, a \textbf{masked multi-head self-attention} mechanism is applied to the sequence of previously generated tokens ($y_1, \ldots, y_{t-1}$). The mask ensures that the prediction of the token $y_t$ can only depend on the preceding tokens, preserving the auto-regressive property.

Second, a \textbf{cross-attention} mechanism allows the decoder to focus on the encoded EEG information. In this sub-layer, the queries ($Q'$) are derived from the output of the decoder's self-attention layer, while the keys ($K'$) and values ($V'$) are derived from the final output of the BART encoder, $H_{enc}$. This is the crucial step where the model synthesizes the neural signal context with the linguistic context.

Finally, the output of the cross-attention sub-layer passes through a position-wise feed-forward network. The output from the final decoder layer is then passed through a linear layer and a softmax function to generate a probability distribution over the entire vocabulary for the next token. This is formally expressed in Equation~\eqref{eq:decoder_prob}.
\begin{equation}
    P(y_t | y_{<t}, H_{enc}) = \text{softmax}(\text{Linear}(\text{decoder\_output}_t))
    \label{eq:decoder_prob}
\end{equation}
This generation process is repeated sequentially until an end-of-sequence token is produced, completing the translation.

\begin{algorithm}[h]
\caption{\textsc{R1Translator}: LSTM–\,BART EEG Decoder}
\label{alg:r1translator}
\begin{algorithmic}[1]
\Require 
    EEG token embeddings $\mathbf{E}\!\in\!\mathbb{R}^{B\times T\times f}$,\;
    attention mask $\mathbf{M}\!\in\!\{0,1\}^{B\times T}$,\;
    inverted mask $\bar{\mathbf{M}}$ (for padding in LSTM),\;
    \textbf{optional} target IDs $\mathbf{Y}$ (only in training) Parameters 
    \begin{tabular}[t]{@{}l@{ }l}
      $f$         & input feature dim (default $840$) \\
      $h$         & LSTM hidden size (default $256$) \\
      $b$         & bidirectional flag ($\in\{0,1\}$) \\
      $L$         & \#\,LSTM layers (default $2$) \\
      $d$         & BART embedding dim (default $1024$)
    \end{tabular}

\vspace{0.4em}
\Statex \textbf{Stage A — EEG Encoder (LSTM)}
\State $\mathbf{H} \gets \textsc{LSTM}(\mathbf{E};\,h,L,b,\texttt{mask}=\bar{\mathbf{M}})$
       \Comment{$\mathbf{H}\!\in\!\mathbb{R}^{B\times T\times h'}$, $h' = h\,(1+b)$}

\Statex \textbf{Stage B — Linear Projection}
\State $\mathbf{Z} \gets \textsc{ReLU}\bigl(W\mathbf{H} + \mathbf{b}\bigr)$
       \Comment{$W\!\in\!\mathbb{R}^{d\times h'}$ aligns with BART’s embed size}

\Statex \textbf{Stage C — Seq2Seq Decoder (Pre-trained BART)}
\If{\textbf{training} (labels $\mathbf{Y}$ are provided)}
    \State $\mathcal{O} \gets \textsc{BART}\bigl(\texttt{inputs\_embeds}=\mathbf{Z},\;
                               \texttt{attention\_mask}=\mathbf{M},\;
                               \texttt{labels}=\mathbf{Y}\bigr)$
    \State \Return loss $\mathcal{L} \leftarrow \mathcal{O}.\texttt{loss}$
\Else \Comment{\textbf{inference / generation}}
    \State $\hat{\mathbf{Y}} \gets 
           \textsc{BART.Generate}\bigl(\texttt{inputs\_embeds}=\mathbf{Z},\;
                                        \texttt{attention\_mask}=\mathbf{M}\bigr)$
    \State \Return decoded tokens $\hat{\mathbf{Y}}$
\EndIf
\end{algorithmic}
\end{algorithm}

\subsection{Two-Step Fine-Tuning Strategy}

The entire training process is divided into two distinct stages, as depicted in Figure \ref{methodology} and formally outlined in Algorithm~\ref{alg:r1_two_step}. This strategy is designed to ensure stable and effective adaptation of the large pretrained language model to the new modality of EEG signals. Let the complete set of trainable parameters in the R1 Translator be denoted by $\theta$. We can partition this set into the parameters of our newly initialized components and the parameters of the pretrained BART model:

\begin{equation}
    \theta = \theta_{LSTM} \cup \theta_{Proj} \cup \theta_{BART}
\end{equation}
The BART parameters, $\theta_{BART}$, are further subdivided into the initial layers targeted for early adaptation ($\theta_{BART-adapter} = \theta_{BART-Emb} \cup \theta_{BART-Enc1}$) and the remaining core pretrained layers ($\theta_{BART-core}$).

\subsubsection{Step 1: Partial Training and Feature Alignment}\label{step:1}

The primary objective of this initial stage is to train the EEG-specific encoder ($\theta_{LSTM}$ and $\theta_{Proj}$) from scratch and gently adapt the input-facing layers of the BART model ($\theta_{BART-adapter}$) to the latent space of the EEG features. During this phase, the core parameters of the BART model, $\theta_{BART-core}$, are kept frozen. This selective training prevents catastrophic forgetting of the powerful linguistic knowledge encoded within the deeper layers of the transformer.

Mathematically, we define the set of trainable parameters for this stage as $\theta_{S1} = \theta_{LSTM} \cup \theta_{Proj} \cup \theta_{BART-adapter}$. The optimization is performed exclusively on this subset, with the gradient updates computed as:
\begin{equation}
    \theta_{S1} \leftarrow \theta_{S1} - \eta_1 \nabla_{\theta_{S1}} L_{CE}(\theta_{S1}; \theta_{BART-core})
    \label{eq:stage1_update}
\end{equation}
where $\eta_1$ is the learning rate for the first stage, and the loss is computed while holding $\theta_{BART-core}$ constant.

\subsubsection{Step 2: End-to-End Fine-Tuning }
Following the initial adaptation, the second stage unfreezes all model parameters to facilitate a deep, co-adaptive integration of all components. This holistic optimization enables the entire architecture, from EEG feature extraction to text generation, to learn in concert.

In this stage, the full set of parameters $\theta$ is made trainable. The model is fine-tuned end-to-end, with parameter updates given by:
\begin{equation}
    \theta \leftarrow \theta - \eta_2 \nabla_{\theta} L_{CE}(\theta)
    \label{eq:stage2_update}
\end{equation}
Typically, a smaller learning rate is employed ($\eta_2 < \eta_1$) to ensure fine-grained adjustments, preserving the knowledge gained in both pretraining and the initial adaptation stage, ultimately leading to optimal performance.

\begin{algorithm}[]
\caption{Two–Stage Training of \textsc{R1 Translator}}
\label{alg:r1_two_step}
\begin{algorithmic}[1]
\Require 
    Training loader $\mathcal{D}_{\mathrm{tr}}$, validation loader $\mathcal{D}_{\mathrm{val}}$,
    max epochs $(N_{1},N_{2})$,            \Comment{$N_{1}$ for \textbf{Step-1}, $N_{2}$ for \textbf{Step-2}}
    learning rates $(\eta_{1},\eta_{2})$,  \Comment{often $\eta_{2}<\eta_{1}$}
    \textsc{R1Translator} model $\theta$ initialized with pre-trained BART
\Statex ---------------------------------------------------------------
\Function{TrainTwoStep}{$\theta$}
    \Statex \textbf{STEP-1: \emph{Feature Adaptation}} (freeze BART except embeddings \& 1st layer)
    \State Freeze $\theta.\texttt{pretrained.encoder.layers}[2\text{:}]$ and decoder $\theta.\texttt{pretrained.decoder}$;
           set their \texttt{requires\_grad}\,$\leftarrow0$
    \State $\textit{opt}\_1 \gets \textsc{SGD}\!\left(\{\theta_i\!\;|\; \texttt{requires\_grad}=1\},\,
                   \text{lr}=\eta_{1}\right)$
    \For{$e\gets0$ \textbf{to} $N_{1}-1$}
        \State $\mathcal{L}_{e}\gets$\Call{EpochTrain}{$\theta,\mathcal{D}_{\mathrm{tr}},\textit{opt}\_1$}
        \State \Call{ValidateAndCheckpoint}{$\theta,\mathcal{D}_{\mathrm{val}},\mathcal{L}_{e},\mathrm{``step1''}$}
    \EndFor
    \Statex \textbf{STEP-2: \emph{Full Fine-tune}}
    \State Unfreeze all params $\theta_i$: \texttt{requires\_grad}$\leftarrow1$
    \State $\textit{opt}\_2 \gets \textsc{SGD}(\theta,\text{lr}=\eta_{2})$
    \For{$e\gets0$ \textbf{to} $N_{2}-1$}
        \State $\mathcal{L}_{e}\gets$\Call{EpochTrain}{$\theta,\mathcal{D}_{\mathrm{tr}},\textit{opt}\_2$}
        \State \Call{ValidateAndCheckpoint}{$\theta,\mathcal{D}_{\mathrm{val}},\mathcal{L}_{e},\mathrm{``step2''}$}
    \EndFor
    \State \Return best checkpoint $\theta^{\ast}$ (lowest val-loss across both steps)
\EndFunction
\vspace{0.2em}
\Function{EpochTrain}{$\theta,\mathcal{D},\textit{opt}$}
    \State $\theta.\texttt{train}()$, $\textit{loss}\leftarrow0$
    \ForAll{batch $(\mathbf{E},\mathbf{M},\bar{\mathbf{M}},\mathbf{Y}) \in \mathcal{D}$}
        \State $\textit{opt}.\texttt{zero\_grad}()$
        \State $\ell \leftarrow \theta(\mathbf{E},\mathbf{M},\bar{\mathbf{M}},\mathbf{Y})$ \Comment{calls Algorithm \ref{alg:r1translator}}
        \State $\ell.\texttt{backward}();\; \textit{opt}.\texttt{step}()$
        \State $\textit{loss} \leftarrow \textit{loss} + \ell$
    \EndFor
    \State \Return $\textit{loss}/|\mathcal{D}|$
\EndFunction
\end{algorithmic}
\end{algorithm}

\subsection{Optimization}

The R1 Translator model was trained to minimize the negative log-likelihood of the target text sequence given the input EEG features. This is achieved using a standard \textbf{cross-entropy loss} function, commonly employed for auto-regressive language modeling tasks. The loss for a single EEG-text pair is defined as:
\begin{equation}
L_{\text{CE}} = -\frac{1}{T} \sum_{t=1}^{T} \log P(y_t | y_{<t}, X_{\text{EEG}})
\label{eq:cross_entropy}
\end{equation}
where $T$ is the total number of tokens in the target text sequence, $y_t$ is the ground-truth token at timestep $t$, $y_{<t}$ represents all ground-truth tokens preceding timestep $t$, $X_{\text{EEG}}$ denotes the processed input EEG features corresponding to the target sequence, and $P(y_t | y_{<t}, X_{\text{EEG}})$ is the probability assigned by the model to the true token $y_t$.

The overall training objective is to find the model parameters $\theta$ that minimize this cross-entropy loss (equation ~\ref{eq:cross_entropy}) across the entire training dataset.

The optimization of the model parameters was performed using Stochastic Gradient Descent (SGD) with momentum. This choice was applied consistently across both stages of the fine-tuning process.
The update rule for SGD with momentum can be generally expressed as:
\begin{gather}
v_k = \mu v_{k-1} + \eta \nabla_{\theta} L(\theta_k) \label{eq:sgd_v} \\
\theta_{k+1} = \theta_k - v_k \label{eq:sgd_theta}
\end{gather}
where $\theta_k$ are the model parameters at iteration $k$, $\nabla_{\theta} L(\theta_k)$ is the gradient of the loss function with respect to the parameters,$\eta$ is the learning rate, $v_k$ is the velocity vector and  $\mu$ is the momentum factor.

Distinct learning rates were utilized for each of the two fine-tuning stages ($\eta_{\text{stage1}}$ and $\eta_{\text{stage2}}$). Furthermore, a step-based learning rate scheduler was employed. This scheduler decayed the learning rate by a factor of $\gamma$ (gamma, set to 0.1) every pre-defined number of epochs (step\_size, set to 20 for Stage 1 and 30 for Stage 2), facilitating more stable convergence during the later phases of training. The specific learning rates and scheduler parameters were empirically determined to yield optimal performance.

\begin{table*}[]
\centering
\caption{Evaluation of EEG-to-Text Decoding Performance on the ZuCo Dataset Using Reading task SR V1, NR V1 and TSR V1: Results with and without Teaching Force (w/tf) During Model Evaluation}
\label{table-v1}
\resizebox{\textwidth}{!}{%
\begin{tabular}{c|cccc|ccc|ccc|ccc}
\hline
\multirow{2}{*}{\textbf{Model}} & \multicolumn{4}{c|}{\textbf{BLEU-N (\%)}} & \multicolumn{3}{c|}{\textbf{ROUGE-1 (\%)}} & \multicolumn{3}{c|}{\textbf{ROUGE-2 (\%)}} & \multicolumn{3}{c}{\textbf{ROUGE-L (\%)}} \\ \cline{2-14} 
                                & N=1       & N=2      & N=3      & N=4     & P            & R            & F            & P            & R            & F            & P            & R            & F           \\ \hline
R1 w/tf                         & 38.62     & 21.41    & 11.65    & 6.15    & \cellcolor{green!30}30.91        & \cellcolor{green!30}25.52        & \cellcolor{green!30}27.79        & \cellcolor{green!30}6.90         & \cellcolor{green!30}6.13         & \cellcolor{green!30}6.45         & \cellcolor{green!30}28.79        & \cellcolor{green!30}23.78        & \cellcolor{green!30}25.90       \\
T5 w/tf                         & \cellcolor{green!30}43.01     & \cellcolor{green!30}24.41    & \cellcolor{green!30}14.15    & \cellcolor{green!30}7.75    & 27.78        & 22.63        & 24.76        & 5.13         & 4.51         & 4.76         & 25.44        & 20.78        & 22.72       \\
Brain w/tf                      & 38.02     & 20.43    & 10.56    & 5.62    & 29.57        & 23.98        & 26.37        & 5.44         & 4.89         & 5.14         & 27.48        & 22.33        & 24.53       \\ \hline
R1                              & 12.16     & 3.22     & \cellcolor{green!30}1.12     & \cellcolor{green!30}0.34    & \cellcolor{green!30}12.01        & 9.51         & \cellcolor{green!30}9.96         & \cellcolor{green!30}1.35         & \cellcolor{green!30}0.84         & \cellcolor{green!30}0.96         & \cellcolor{green!30}10.58        & 8.41         & \cellcolor{green!30}8.75        \\
T5                              & 13.98     & \cellcolor{green!30}3.53     & 0.92     & 0.27    & 10.08        & 9.27         & 9.03         & 0.36         & 0.28         & 0.30         & 8.39         & 7.99         & 7.62        \\
Brain                           & \cellcolor{green!30}14.10     & 2.62     & 0.66     & 0.25    & 9.22         & \cellcolor{green!30}11.91        & 9.75         & 0.33         & 0.39         & 0.34         & 7.76         & \cellcolor{green!30}10.34        & 8.31        \\ \hline
\end{tabular}%
}
\end{table*}

\begin{figure*}[htbp]
    \centering
    \begin{subfigure}{0.33\textwidth}
        \centering
        \includegraphics[width=\textwidth]{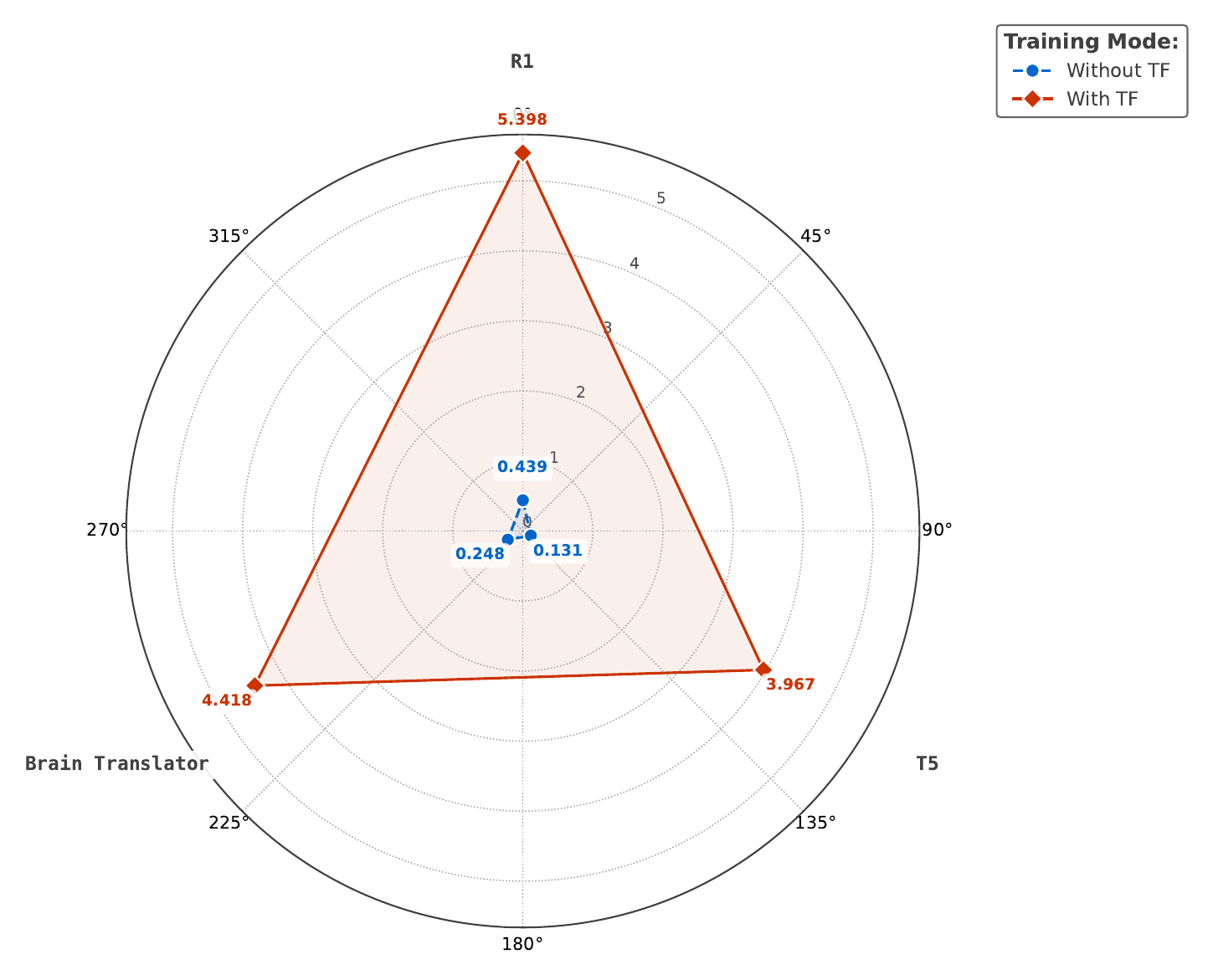} 
        \caption{SacreBLEU scores }
    \end{subfigure}%
    \begin{subfigure}{0.33\textwidth}
        \centering
        \includegraphics[width=\textwidth]{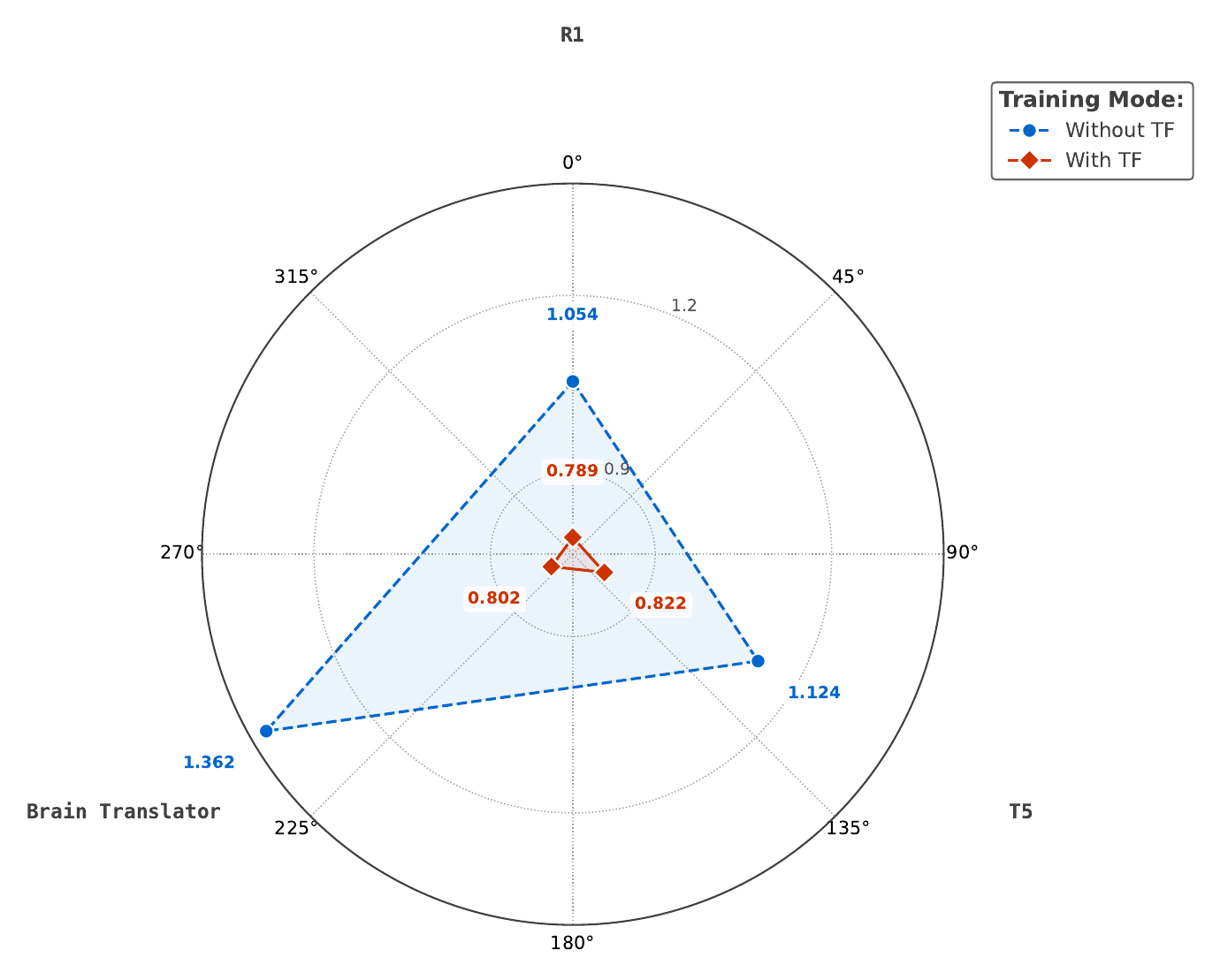} 
        \caption{Word Error Rate (WER)}
    \end{subfigure}%
    \begin{subfigure}{0.33\textwidth}
        \centering
        \includegraphics[width=\textwidth]{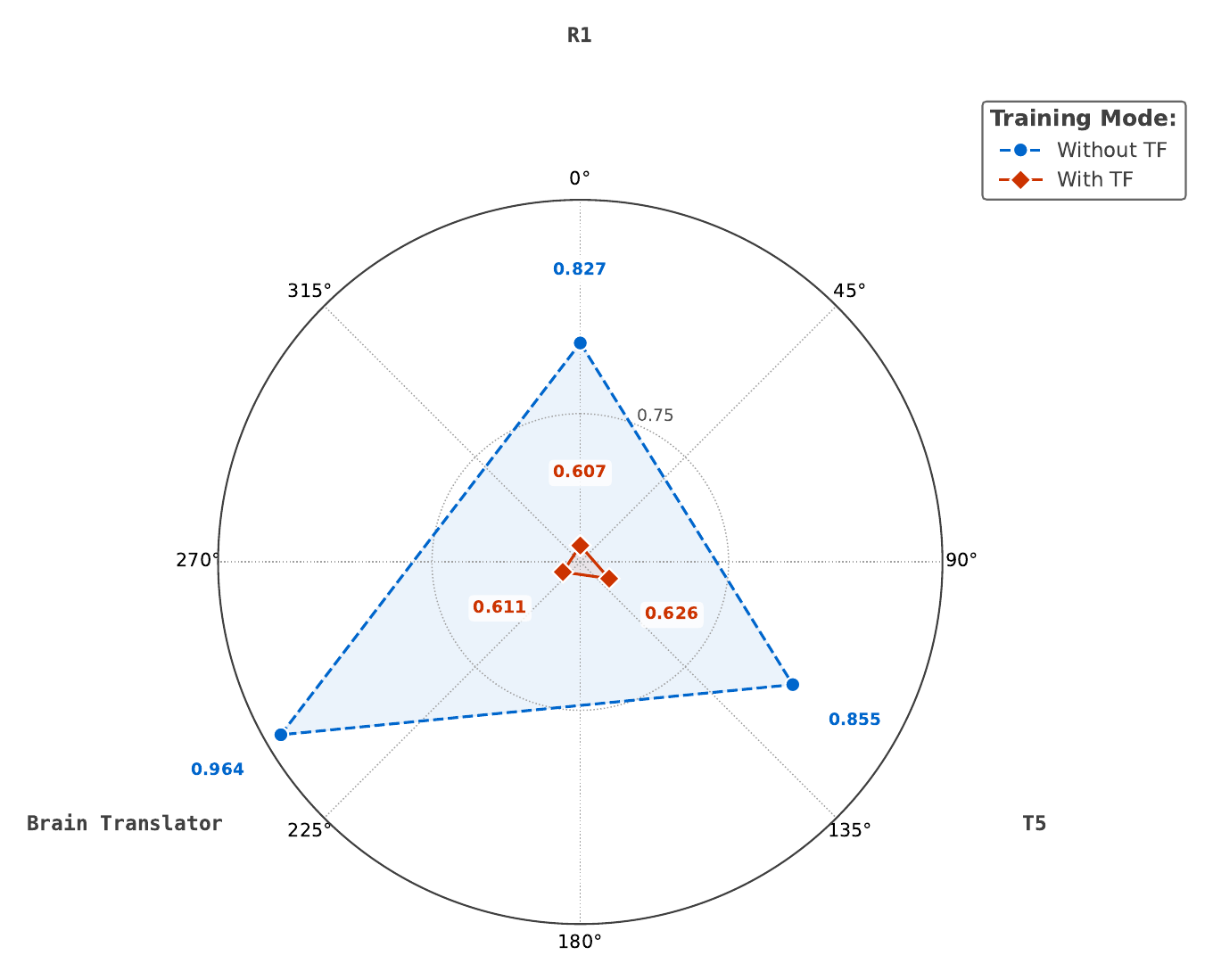} 
        \caption{ Character Error Rate (CER) }
    \end{subfigure}
    \caption{Evaluation of SacreBLEU, Word Error Rate (WER), and Character Error Rate (CER) with and without teaching force (w/tf) for EEG-to-text decoding on the ZuCo dataset, across Reading Task versions SR V1, NR V1, and TSR V1.}
    \label{fig_v1}
\end{figure*}

\begin{table*}[]
\centering
\caption{Evaluation of EEG-to-Text Decoding Performance on the ZuCo Dataset Using Reading Task NR V2 and TSR V2: Results with and without Teaching Force (w/tf) During Model Evaluation}
\label{table-V2}
\resizebox{\textwidth}{!}{%
\begin{tabular}{c|cccc|ccc|ccc|ccc}
\hline
\multirow{2}{*}{\textbf{Model}} & \multicolumn{4}{c|}{\textbf{BLEU-N (\%)}} & \multicolumn{3}{c|}{\textbf{ROUGE-1 (\%)}} & \multicolumn{3}{c|}{\textbf{ROUGE-2 (\%)}} & \multicolumn{3}{c}{\textbf{ROUGE-L (\%)}} \\ \cline{2-14} 
                                & N=1      & N=2      & N=3      & N=4      & P            & R            & F            & P            & R            & F            & P            & R            & F           \\ \hline
R1 w/tf                         & 45.42    & 27.53    & 16.86    & 10.40    & \cellcolor{green!30}41.08        & \cellcolor{green!30}33.87        & \cellcolor{green!30}36.99        & \cellcolor{green!30}12.35        & \cellcolor{green!30}11.34        & \cellcolor{green!30}11.81        & \cellcolor{green!30}38.88        & \cellcolor{green!30}32.01        &\cellcolor{green!30}34.97       \\
T5 w/tf                         & \cellcolor{green!30}47.59    & \cellcolor{green!30}29.08    & \cellcolor{green!30}17.84    & \cellcolor{green!30}11.08    & 38.03        & 31.96        & 34.52        & 10.62        & 9.76         & 10.13        & 36.19        & 30.43        & 32.86       \\
Brain w/tf                      & 0.00     & 0.00     & 0.00     & 0.00     & 0.00         & 0.00         & 0.00         & 0.00         & 0.00         & 0.00         & 0.00         & 0.00         & 0.00        \\ \hline
R1                              & 16.59    & 3.59     & 1.40     & 0.64     & \cellcolor{green!30}18.17        & 16.76        & \cellcolor{green!30}16.59        & 1.26         & 0.96         & 0.97         & 14.97        & 13.93        & 13.70       \\
T5                              & \cellcolor{green!30}18.97    & \cellcolor{green!30}7.28     & \cellcolor{green!30}3.06     & \cellcolor{green!30}1.45     & 17.92        & 16.90        & 16.47        & \cellcolor{green!30}2.17         & \cellcolor{green!30}2.24         & \cellcolor{green!30}2.01         & \cellcolor{green!30}15.39        & 14.70        & \cellcolor{green!30}14.22       \\
Brain                           & 13.82    & 1.61     & 0.21     & 0.00     & 10.30        & \cellcolor{green!30}24.42        & 14.02        & 0.34         & 0.14         & 0.19         & 7.44         & \cellcolor{green!30}18.08        & 10.20       \\ \hline
\end{tabular}%
}
\end{table*}

\begin{figure*}[htbp]
    \centering
    \begin{subfigure}{0.33\textwidth}
        \centering
        \includegraphics[width=\textwidth]{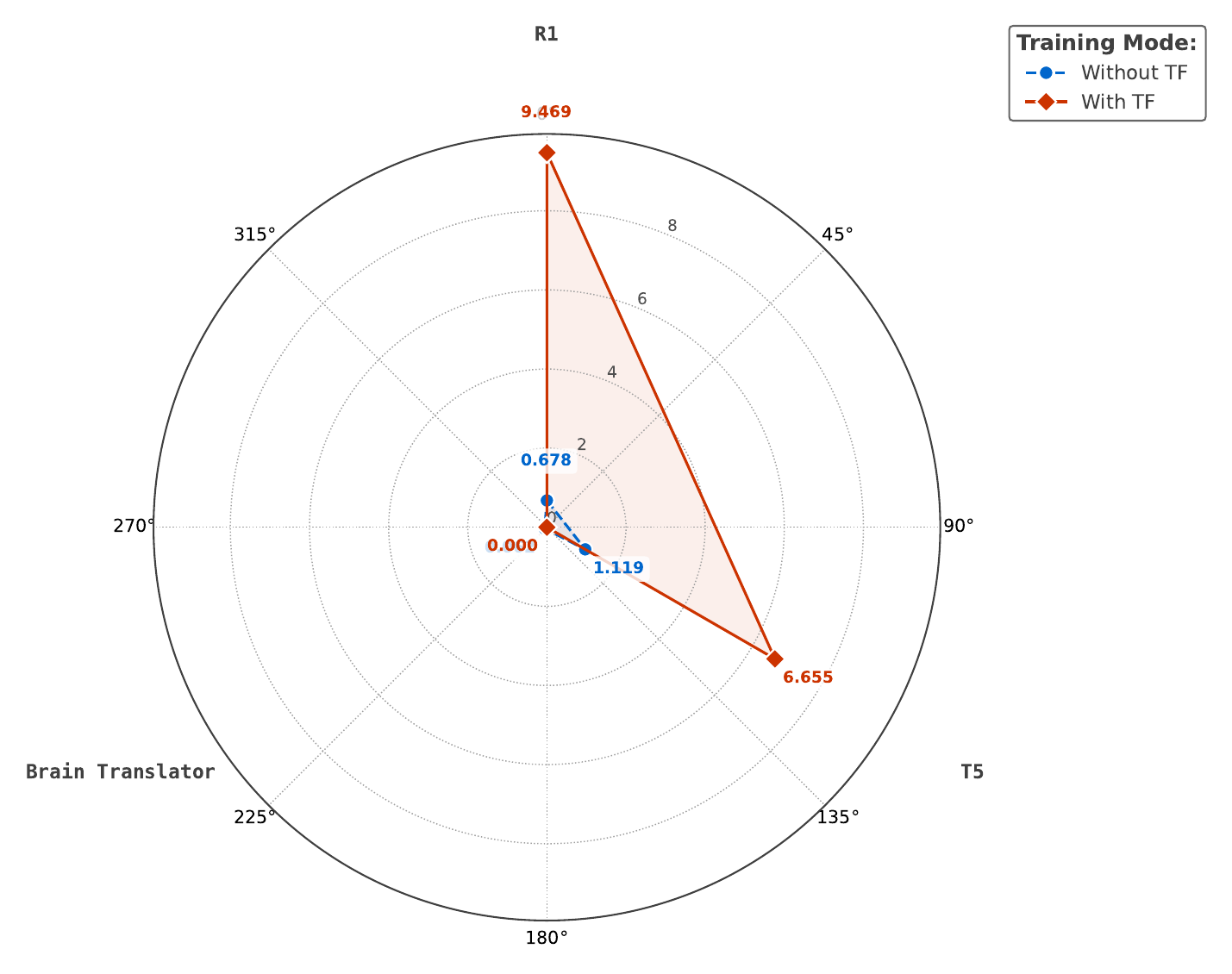} 
        \caption{SacreBLEU scores }
    \end{subfigure}%
    \begin{subfigure}{0.33\textwidth}
        \centering
        \includegraphics[width=\textwidth]{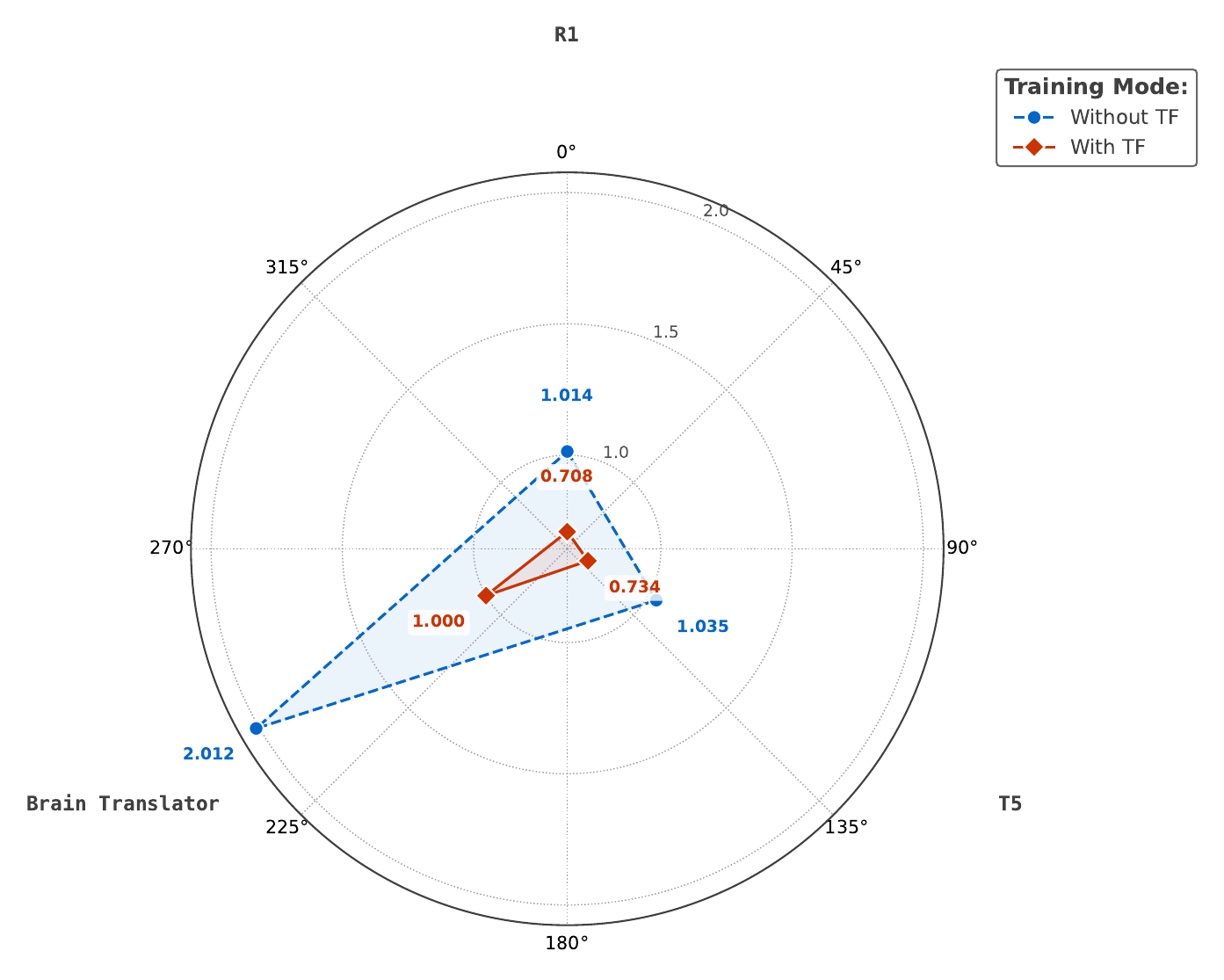} 
        \caption{Word Error Rate (WER)}
    \end{subfigure}%
    \begin{subfigure}{0.33\textwidth}
        \centering
        \includegraphics[width=\textwidth]{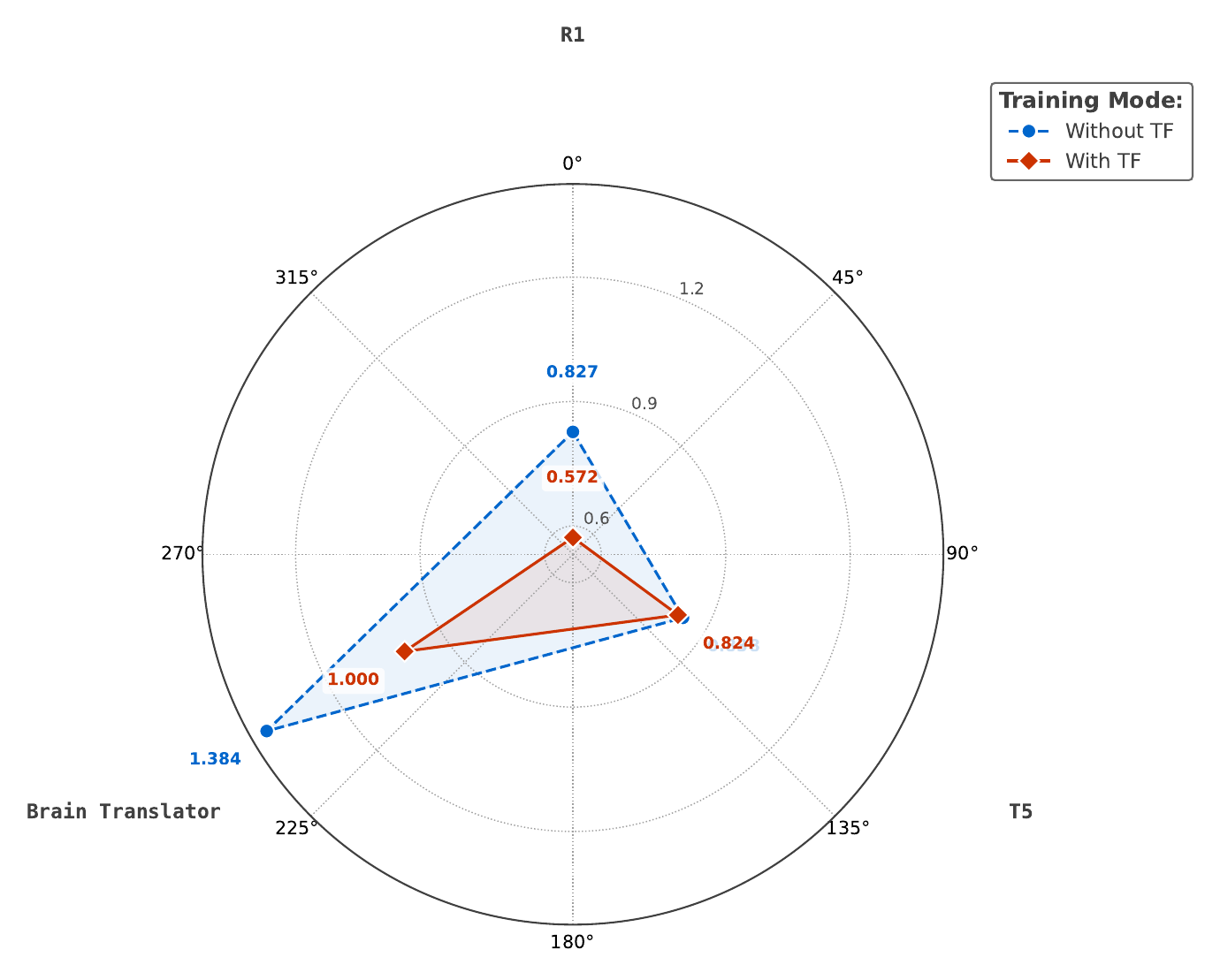} 
        \caption{ Character Error Rate (CER) }
    \end{subfigure}
    \caption{Evaluation of SacreBLEU, Word Error Rate (WER), and Character Error Rate (CER) with and without teaching force (w/tf) for EEG-to-text decoding on the ZuCo dataset, across Reading Task versions NR V2 and TSR V2.}
    \label{fig-v2}
\end{figure*}

\section{Implementation Details}
\label{implementation}
We implemented our R1 Translator model within the PyTorch framework for our EEG-to-text translation, utilizing pretrained components from the Hugging Face Transformers library (e.g., BART, T5). The core methodology involved processing ZuCo dataset EEG signals, specifically features from multiple frequency bands (8 bands, 105 channels each), which were normalized per word. These EEG features were then fed into a custom encoder, which projects the EEG representations into the input embedding space of the main pretrained transformer model. Training was conducted on an NVIDIA A40-48Q GPU with 48GB of memory, utilizing CUDA version 12.4, through a two-step fine-tuning approach for R1 Translator, as indicated by our evaluation configurations: an initial 20 epochs followed by 30 epochs, both typically with learning rates around 2e-5, an SGD optimizer, and a batch size of 32, using EEG data as input. To assess the quality of the generated text from unseen EEG recordings, we benchmarked performance using a suite of standard metrics, which are formally defined below.

\subsection{Evaluation Metrics}

\paragraph{BLEU}
The Bilingual Evaluation Understudy (BLEU) score measures the n-gram precision between the generated text and a set of reference texts. It is defined as:
\begin{equation}
    \text{BLEU} = \text{BP} \cdot \exp\left(\sum_{n=1}^{N} w_n \log p_n\right)
\end{equation}
where $p_n$ is the modified n-gram precision for n-grams of size $n$, $w_n$ are weights (typically uniform), and BP is a brevity penalty to penalize generated texts that are too short. The brevity penalty is calculated as:
\begin{equation}
    \text{BP} = 
    \begin{cases} 
      1 & \text{if } c > r \\
      e^{(1 - r/c)} & \text{if } c \le r 
    \end{cases}
\end{equation}
where $c$ is the length of the candidate text and $r$ is the effective reference length. We also report SacreBLEU, which is a standardized implementation of BLEU that uses a common tokenization scheme to ensure comparability across different studies.

\paragraph{ROUGE-L}
The Recall-Oriented Understudy for Gisting Evaluation (ROUGE-L) metric is based on the Longest Common Subsequence (LCS). It measures the F1-score of the LCS between a candidate text $Y$ and a reference text $X$. The recall ($R_{lcs}$) and precision ($P_{lcs}$) are first computed:
\begin{equation}
    R_{lcs} = \frac{\text{LCS}(X, Y)}{m} \quad \text{and} \quad P_{lcs} = \frac{\text{LCS}(X, Y)}{n}
\end{equation}
where $m$ and $n$ are the lengths of the reference and candidate texts, respectively. The final ROUGE-L score is the F-score:
\begin{equation}
    \text{ROUGE-L} = \frac{(1 + \beta^2) R_{lcs} P_{lcs}}{R_{lcs} + \beta^2 P_{lcs}}
\end{equation}
where $\beta$ is set to favor recall over precision.

\paragraph{WER and CER}
Word Error Rate (WER) and Character Error Rate (CER) are metrics derived from the Levenshtein distance, measuring the number of edits required to transform the generated text into the reference text. WER operates at the word level and is defined as:
\begin{equation}
    \text{WER} = \frac{S + D + I}{N}
\end{equation}
where $S$, $D$, and $I$ are the number of substitutions, deletions, and insertions, respectively, and $N$ is the total number of words in the reference text. CER is computed identically but at the character level, providing a more granular measure of textual similarity. For both metrics, a lower score indicates better performance.

\section{Results}
\label{results}

In our experiments, we explored three different approaches. First, we conducted experiments using the ZuCo V1 dataset, followed by a second round using the ZuCo V2 dataset. Finally, we combined both datasets to perform a third experiment. For each of these steps, we evaluated the performance of two established models—T5 and Brain Translator—alongside our proposed model, R1 Translator. Notably, R1 Translator outperformed the other models, achieving SOTA results across all experiments.

\begin{table*}[h]
\centering
\caption{Evaluation of EEG-to-Text Decoding Performance on the ZuCo Dataset Using Reading Task SR V1.0, NR V1.0, TSR V1, and TSR V2: Results with and without Teaching Force (w/tf) During Model Evaluation}
\label{Tab-v1-v2}
\resizebox{\textwidth}{!}{%
\begin{tabular}{c|cccc|ccc|ccc|ccc}
\hline
\multirow{2}{*}{\textbf{Model}} & \multicolumn{4}{c|}{\textbf{BLEU-N (\%)}} & \multicolumn{3}{c|}{\textbf{ROUGE-1 (\%)}} & \multicolumn{3}{c|}{\textbf{ROUGE-2 (\%)}} & \multicolumn{3}{c}{\textbf{ROUGE-L (\%)}} \\ \cline{2-14} 
                                & N=1      & N=2      & N=3      & N=4     & P            & R            & F           & P            & R            & F           & P            & R            & F           \\ \hline \hline
R1 w/tf                         & 44.44    & 26.63    & 16.05    & 9.93    & \cellcolor{green!30}38.00        & \cellcolor{green!30}31.75        & \cellcolor{green!30}34.47       & \cellcolor{green!30}10.95        & \cellcolor{green!30}10.07        & \cellcolor{green!30}10.47       & \cellcolor{green!30}35.81        & \cellcolor{green!30}29.95        & \cellcolor{green!30}32.51       \\ 
T5 w/tf                         & \cellcolor{green!30}46.42    & \cellcolor{green!30}28.33    & \cellcolor{green!30}17.34    & \cellcolor{green!30}10.56   & 34.89        & 28.93        & 31.44       & 9.52         & 8.41         & 8.86        & 32.87        & 27.32        & 29.67       \\
Brain w/tf                      & 40.95    & 23.24    & 13.21    & 7.55    & 35.69        & 29.04        & 31.89       & 9.51         & 8.62         & 9.03        & 34.03        & 27.63        & 30.38       \\ \hline \hline

R1                              & \cellcolor{green!30}18.22    & 4.37     & 1.42     & 0.56    & 15.62        & 15.93        & \cellcolor{green!30}15.31       & 1.07         & 1.01         & 0.98        & 12.77        & 13.15        & 12.55       \\ 

T5                              & 17.9     & \cellcolor{green!30}6.71     & \cellcolor{green!30}2.72     & \cellcolor{green!30}1.25    & \cellcolor{green!30}16.39        & \cellcolor{green!30}16.00        & \cellcolor{green!30}15.31       & \cellcolor{green!30}1.81         & \cellcolor{green!30}1.86         & \cellcolor{green!30}1.64        & \cellcolor{green!30}13.85        & \cellcolor{green!30}13.72        & \cellcolor{green!30}13.01       \\ 

Brain                           & 14.28    & 3.92     & 1.49     & 0.66    & 15.87        & 13.51        & 13.83       & 1.68         & 1.48         & 1.42        & 13.69        & 11.75        & 11.97       \\ \hline
\end{tabular}%
}
\end{table*}

\begin{figure*}[htbp]
    \centering
    \begin{subfigure}{0.33\textwidth}
        \centering
        \includegraphics[width=\textwidth]{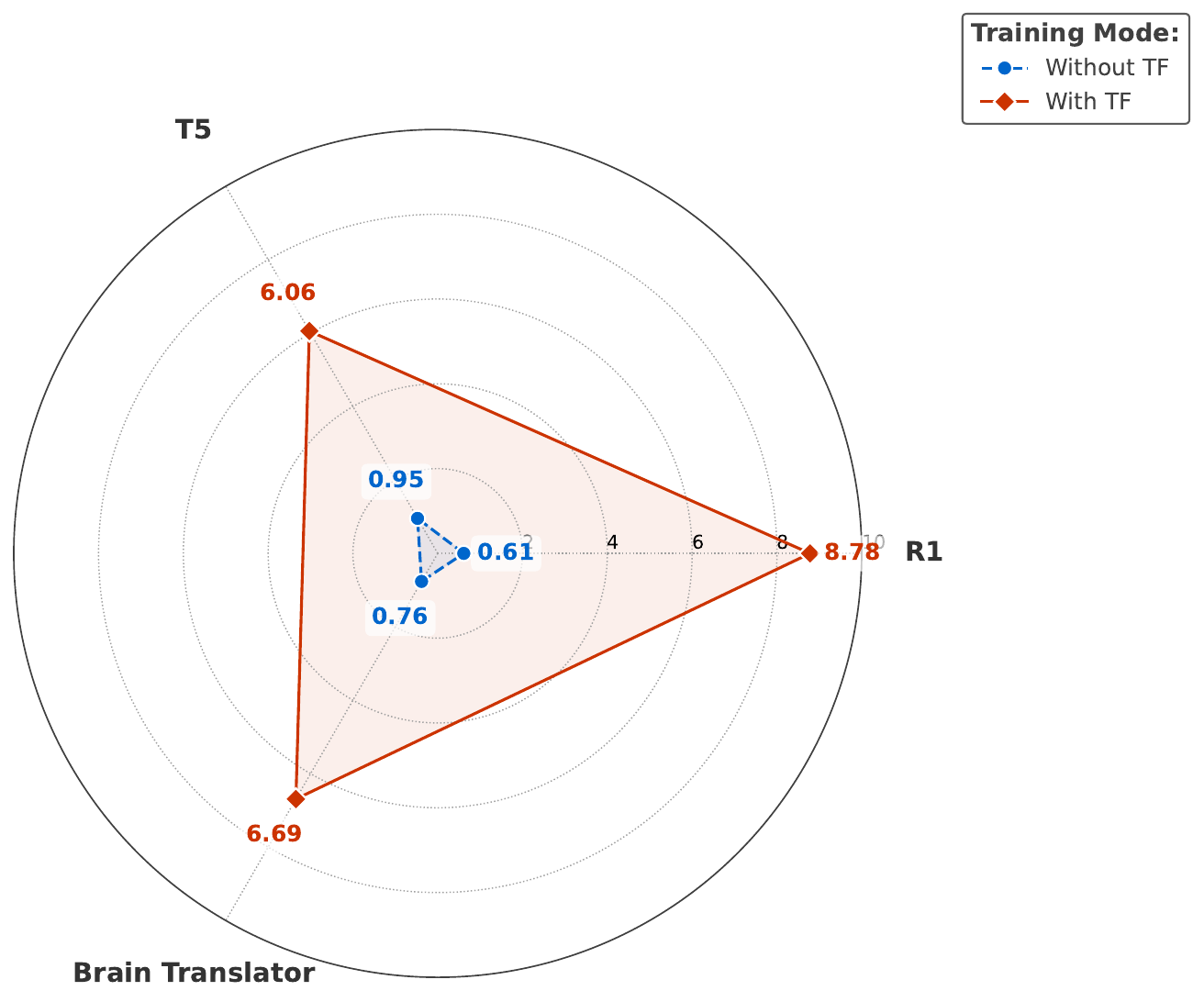} 
        \caption{SacreBLEU scores }
    \end{subfigure}%
    \begin{subfigure}{0.33\textwidth}
        \centering
        \includegraphics[width=\textwidth]{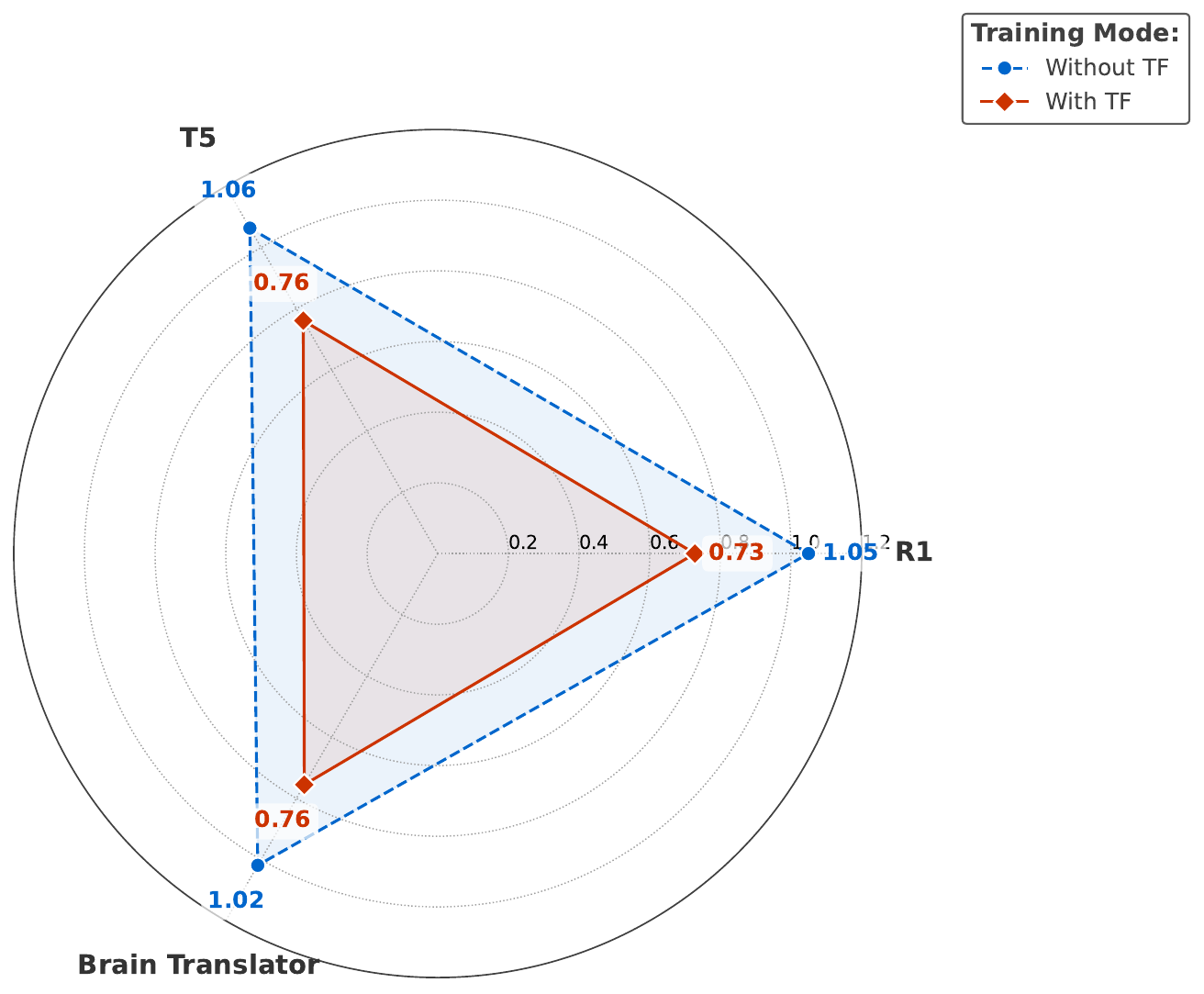} 
        \caption{Word Error Rate (WER)}
    \end{subfigure}%
    \begin{subfigure}{0.33\textwidth}
        \centering
        \includegraphics[width=\textwidth]{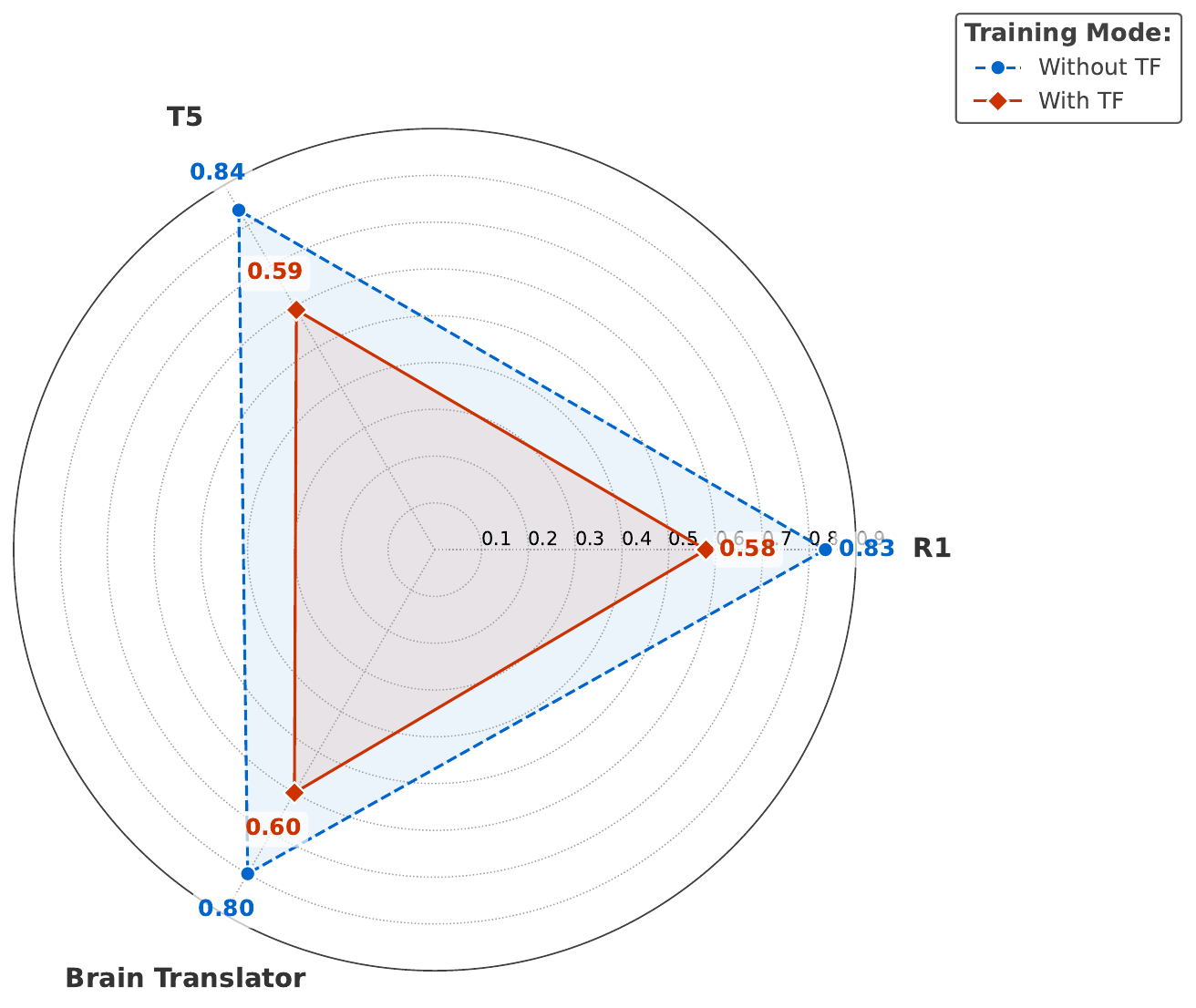} 
        \caption{ Character Error Rate (CER) }
    \end{subfigure}
    \caption{Evaluation of SacreBLEU, Word Error Rate (WER), and Character Error Rate (CER) with and without teaching force (w/tf) for EEG-to-text decoding on the ZuCo dataset, across Reading Task versions SR V1.0, NR V1.0, TSR V1, and TSR V2.}
    \label{fig_v1_v2}
\end{figure*}

\begin{table*}[]
\centering
\caption{This table compares the performance of two translation models, T5 Translator and R1 Translator, by decoding EEG unseen wave data. The ground truth translations are provided, with matching words highlighted in \textbf{bold} for easy comparison. The table illustrates how accurately each model replicates key elements of the original text. }
\label{comparison_table}
\resizebox{\textwidth}{!}{%
\begin{tabular}{@{}ll@{}}
\toprule
Ground Truth          & \begin{tabular}[c]{@{}l@{}}Warm Water Under a Red \textbf{Bridge} \textbf{is a} quirky and poignant Japanese \textbf{film} that explores \textbf{the} fascinating connections \\ \textbf{between} women, water, nature, \textbf{and} sexuality.\end{tabular}                                                                           \\ 
R1 Translator         & \begin{tabular}[c]{@{}l@{}}altth is the \textbf{Bridge} Sun \textbf{is a} film, funny film \textbf{film} about is \textbf{the} relationship relationship \textbf{between} the and \\ science, \textbf{and}, and the.\end{tabular}                                                                                                 \\
T5 Translator \cite{intro6} & \begin{tabular}[c]{@{}l@{}}The,s thea Storm Sky, a \textbf{film}, ant tale- that iss \textbf{the} relationship world \textbf{between} the and the, \textbf{and} and and theity.\end{tabular}                                                                                                   \\ \midrule
Ground Truth          & An amateurish, quasi-improvised acting exercise shot on ugly digital \textbf{video.}                                                                            \\ 
R1 Translator         & interesting filmmaker, un-religiousprovised film performance, through a, \textbf{video.}                                                                                                 \\
T5 Translator \cite{intro6} & ’He American radio, -svised  career, to the, filmtap                                                                                                  \\ \midrule
Ground Truth          & It's not a particularly \textbf{good} film, \textbf{but} neither \textbf{is it} a monsterous \textbf{one}.                                                                            \\ 
R1 Translator         & 's a a bad \textbf{good} movie, \textbf{but} it \textbf{is it} bad bad. \textbf{one}.                                                                                                 \\
T5 Translator \cite{intro6} & ’s a a bad \textbf{good} movie, \textbf{but} it \textbf{is it} bad bad. \textbf{one}                                                                                                  \\ \midrule
Ground Truth          & \begin{tabular}[c]{@{}l@{}}Everything its title \textbf{implies}, a standard-\textbf{issue} crime \textbf{drama} spat out from the Tinseltown assembly \textbf{line}.\end{tabular} \\
R1 Translator         & \begin{tabular}[c]{@{}l@{}}about predecessor \textbf{implies} is including man,\textbf{issue}, \textbf{drama}. between of the sameseltown set \textbf{line}.\end{tabular}          \\
T5 Translator \cite{intro6} & \begin{tabular}[c]{@{}l@{}}about predecessor \textbf{implies} is and movie,\textbf{issue}, \textbf{drama}. between of the depthsseltown set \textbf{line}.\end{tabular}             \\ \midrule
\end{tabular}%
}
\end{table*}

\subsection{Evaluation on ZuCo V1 Dataset}

Table \ref{table-v1} presents the evaluation of EEG-to-Text decoding performance on the ZuCo V1 dataset using three models: R1 Translator, T5, and Brain, with and without Teaching Force (w/tf). R1 Translator consistently outperforms both T5 and Brain across all metrics, with the highest BLEU-N scores in both settings, especially excelling in BLEU-N at N=1 (38.62\%) and N=2 (21.41\%) when using Teaching Force. Additionally, R1 shows superior performance in the ROUGE-1, ROUGE-2, and ROUGE-L metrics, particularly in Precision (P), Recall (R), and F-Score (F), all highlighted in green, indicating its strong ability to produce accurate translations. Without Teaching Force, R1 Translator still maintains its lead, especially in BLEU-N at N=3 (1.12\%) and N=4 (0.34\%) as well as in ROUGE scores, confirming its robustness across both scenarios. While T5 and Brain perform reasonably well, especially in some ROUGE metrics, R1 Translator’s consistent superiority in both BLEU and ROUGE evaluations highlights its effectiveness in EEG-to-text decoding on the ZuCo V1 dataset.

Figure \ref{fig_v1} provide a detailed comparison of the performance of three models across three key evaluation metrics: sacreBLEU, WER, and CER. In terms of sacreBLEU, R1 Translator shows significant superiority over the other models, achieving an 8\% higher score than T5 and 10\% higher than Brain Translator when Teaching Force is applied. For WER, R1 Translator outperforms T5 by approximately 5\% with Teaching Force, and by a larger margin of 7\% compared to Brain Translator. In CER, R1 Translator delivers an 8\% improvement over T5 and a 9\% improvement over Brain Translator when Teaching Force is applied, further highlighting its superior decoding accuracy. Even without Teaching Force, R1 Translator maintains its lead, achieving 3\% higher performance than T5 and 5\% higher than Brain for CER, and 6\% higher in WER.

\subsection{Evaluation on ZuCo V2 Dataset}

Table \ref{table-V2} presents the evaluation of the decoding performance on the ZuCo V2 dataset. R1 Translator demonstrates superior performance across all evaluation metrics, particularly in BLEU-N and ROUGE scores, both with and without Teaching Force. When Teaching Force is applied, R1 Translator achieves the highest BLEU-N scores, notably at N=1 (45.42\%) and N=2 (27.53\%), and leads in ROUGE-1 (P: 41.08\%, R: 33.87\%, F: 36.99\%) and ROUGE-L (P: 38.88\%, R: 32.01\%, F: 34.97\%) metrics. Even without Teaching Force, R1 maintains its lead, delivering strong results in BLEU-N, with the best performance in N=1 (16.59\%) and N=2 (3.59\%), and excelling in ROUGE-1 (P: 18.17\%) and ROUGE-L (P: 14.97\%). In contrast, T5 also performs well, particularly in BLEU-N, with scores close to R1, but lags behind in ROUGE metrics, showing the gap between R1 and T5 in translation quality and accuracy.

In Figure \ref{fig-v2}, R1 Translator consistently outperforms the other models in all metrics, achieving the highest scores in sacreBLEU, with a 10\% improvement over T5 and nearly 100\% better than Brain Translator when Teaching Force is applied. Similarly, for WER and CER, R1 Translator shows substantial improvements, outperforming T5 by approximately 5\% in WER and 8\% in CER with Teaching Force, and continuing to lead even without Teaching Force. While T5 shows competitive performance, especially in BLEU-N, its ROUGE scores lag behind those of R1 Translator.

\subsection{Evaluation on ZuCo V1 and V2 Dataset}

Table \ref{Tab-v1-v2} presents the evaluation of EEG-to-Text decoding performance on the combined ZuCo V1 and V2 datasets. R1 Translator again stands out as the top performer across all metrics, both with and without Teaching Force. With Teaching Force, R1 achieves the highest BLEU-N scores, notably at N=1 (44.44\%) and N=2 (26.63\%), as well as the best ROUGE-1 (P: 38.00\%, R: 31.75\%, F: 34.47\%) and ROUGE-L (P: 35.81\%, R: 29.95\%, F: 32.51\%) results. Even without Teaching Force, R1 Translator maintains its superiority, especially in BLEU-N at N=1 (18.22\%) and N=2 (4.37\%), and in ROUGE-1 (P: 15.62\%) and ROUGE-L (P: 12.77\%), demonstrating its consistency across both configurations. T5 shows competitive performance, particularly in BLEU-N, but lags behind R1 in ROUGE metrics.

In Figure \ref{fig_v1_v2} R1 Translator consistently outperforms both T5 and Brain Translator across all metrics, particularly excelling in sacreBLEU, where it achieves a substantial 8.6\% improvement over T5 and 9.5\% over Brain with Teaching Force. In WER, R1 shows 5\% better performance than T5 and 8\% over Brain with Teaching Force, and maintains similar advantages without Teaching Force. For CER, R1 Translator delivers 10\% higher accuracy compared to T5 and 14\% compared to Brain when Teaching Force is applied. 

\subsection{Generated Text Samples}

Table \ref{comparison_table} compares the performance of T5 Translator and R1 Translator in decoding EEG unseen wave data, with ground truth translations provided for reference. R1 Translator consistently outperforms T5 in terms of accurately replicating key elements of the original text. Notably, R1 does a better job at maintaining meaningful and accurate translations, particularly in terms of highlighting important words such as "Bridge," "film," and "between," which are correctly captured in R1's translations, while T5 occasionally produces garbled or irrelevant output. For example, R1 successfully maintains the coherence and intent of phrases like "good film" and "implies," which are critical to the meaning, whereas T5's translations show more disruption and loss of context. R1 Translator also shows a clearer understanding of sentence structure and meaning, especially in phrases like "but neither is it a monstrous one," where T5's output becomes incomprehensible. These improvements in R1's accuracy suggest that it is more adept at maintaining the integrity of the original text compared to T5, as highlighted in the ground truth.

\section{Limitations}
\label{limitations}

While the R1 Translator model shows promising results, several limitations need to be addressed. One key limitation is the reliance on the ZuCo dataset, which, although valuable, may not fully capture the diversity of real-world EEG data, as it focuses primarily on movie reviews and Wikipedia articles. This makes it harder to generalize the findings to other types of text or languages. Furthermore, the quality of the EEG signals used in the study plays a significant role in performance, and noise or artifacts in the data could impact the precision of the decoding process, especially in more natural settings of the real world. Although R1 Translator outperforms existing models in many aspects, it still struggles with handling more complex or nuanced sentences, as seen in some of the BLEU-N results. Moreover, the model has not yet been tested in real-time or clinical environments, which limits its practical applicability.

\section{Conclusion \& Future Work}
\label{conclusion}

In this research article, we have addressed the limitations of previous research articles. To overcome the limitations of previous work, we have proposed a model called R1 Translator. This model integrates an LSTM network with both BERT encoder and decoder components, offering a more robust and efficient approach for EEG-to-text decoding. We compared the performance of R1 Translator with the recent research article \cite{intro6}, showing significant improvements in generating meaningful sentences from EEG signals. For our experiments, we utilized the ZuCo dataset versions 1.0 and 2.0, to evaluate the model's performance. To assess the robustness of R1 Translator, we conducted the experiments in three different ways: first, by combining all of ZuCo V1's datasets (SR v1.0, NR v1.0, and TSR v1.0); second, by combining ZuCo V1's TSR v2.0 and NR v2.0; and third, by merging all datasets from both ZuCo V1 and V2. In all three experiment setups, R1 Translator consistently generated coherent and contextually accurate sentences from the EEG signals, demonstrating its reliability and effectiveness in real-world applications.

Looking ahead, our future work will proceed in two primary directions. First, to address the limitations of relying on existing datasets, we plan to collect a larger and more diverse corpus of EEG data from a wider range of participants and tasks. This new dataset will be instrumental in further training the R1 Translator, with the primary goal of enhancing its robustness and generalizability across different subjects and linguistic contexts. Second, a significant next step involves transitioning from offline analysis to real-time application. To achieve this, we will integrate the R1 Translator with our previously designed Linear Diophantine equation based P2P Two-level Hierarchical (LDEPTH) based P2P architecture \cite{p2p}. This integration aims to develop a decentralized system capable of translating EEG signals into text instantaneously, paving the way for practical, real-time brain-computer communication interfaces.

\end{document}